\documentclass[conference]{IEEEtran}
\IEEEoverridecommandlockouts
\usepackage{cite}
\usepackage{url}
\usepackage{amsmath,amssymb,amsfonts,amsthm}
\usepackage{algorithmic}
\usepackage{graphicx}
\usepackage{textcomp}
\usepackage{xcolor}
\usepackage{subfigure}
\usepackage{algorithm, algorithmic}

\usepackage[subnum]{cases}
\newtheorem{theorem}{Theorem}

\newtheorem{remark}{Remark}

\newtheorem{assumption}{Assumption} 
\newtheorem{corollary}{Corollary}
\newcommand\numberthis{\addtocounter{equation}{1}\tag{\theequation}}

\graphicspath{ {./Figures/} }
\IEEEoverridecommandlockouts
\usepackage{tikz}
\usepackage{textcomp}
\usepackage{hyperref}
\usepackage{lipsum}
\usepackage[capitalize]{cleveref}

\newcommand\copyrighttext{%
  \footnotesize \textcopyright 2021 IEEE. Personal use of this material is permitted. Permission from IEEE must be obtained for all other uses, in any current or future media, including reprinting/republishing this material for advertising or promotional purposes, creating new collective works, for resale or redistribution to servers or lists, or reuse of any copyrighted component of this work in other works. Accepted to be Published in: IEEE GLOBECOM 2021 Conference Proceedings, December 7-11, 2021, Madrid,Spain//Hybrid:In-Person and Virtual Conference.}

\newcommand\copyrightnotice{%
\begin{tikzpicture}[remember picture,overlay]
\node[anchor=south,yshift=10pt] at (current page.south) {\fbox{\parbox{\dimexpr\textwidth-\fboxsep-\fboxrule\relax}{\copyrighttext}}};
\end{tikzpicture}%
}

\begin{document}

\title{
To Talk or to Work: Delay Efficient Federated Learning over Mobile Edge Devices
\author{
\IEEEauthorblockN{Pavana Prakash\IEEEauthorrefmark{1}, Jiahao Ding\IEEEauthorrefmark{1}, Maoqiang Wu\IEEEauthorrefmark{2}, Minglei Shu\IEEEauthorrefmark{3}, Rong Yu\IEEEauthorrefmark{2}, and Miao Pan\IEEEauthorrefmark{1}}
\IEEEauthorblockA{\IEEEauthorrefmark{1}Department of Electrical and Computer Engineering, University of Houston, Houston, TX 77204}
\IEEEauthorblockA{\IEEEauthorrefmark{2}School of Automation, Guangdong University of Technology, Guangzhou, China}
\IEEEauthorblockA{{\IEEEauthorrefmark{3}Shandong Artificial Intelligence Institute, Qilu University of Technology (Shandong Academy of Sciences), Jinan, China}}
}
}

\maketitle

\copyrightnotice

\begin{abstract}
Federated learning (FL), an emerging distributed machine learning paradigm, in conflux with edge computing is a promising area with novel applications over mobile edge devices. In FL, since mobile devices collaborate to train a model based on their own data under the coordination of a central server by sharing just the model updates, training data is maintained private. However, without the central availability of data, computing nodes need to communicate the model updates often to attain convergence. Hence, the local computation time to create local model updates along with the time taken for transmitting them to and from the server result in a delay in the overall time. Furthermore, unreliable network connections may obstruct an efficient communication of these updates. To address these, in this paper, we propose a delay-efficient FL mechanism that reduces the overall time (consisting of both the computation and communication latencies) and communication rounds required for the model to converge. Exploring the impact of various parameters contributing to delay, we seek to balance the trade-off between wireless communication (to talk) and local computation (to work). We formulate a relation with overall time as an optimization problem and demonstrate the efficacy of our approach through extensive simulations.
\end{abstract}

\section{Introduction}
Machine learning together with increased capabilities in mobile devices have led to a tremendous rise in the number of smart mobile devices and data generated at the edge network. About $80$ billion devices are predicted to be connected to the Internet by 2025~\cite{reinsel2018digitization}. Hence, computing networks are witnessing a paradigm shift from conventional cloud computing setting, by moving closer to the edge where data is produced, namely multi-access edge computing (MEC). However, utilizing centralized machine learning algorithms at the response-accelerated MEC is inefficient, since uploading and storing bulk data causes a large storage and communication bottleneck. Therefore, federated learning (FL) was introduced to solve these challenges where mobile devices jointly train a shared global model in a decentralized manner~\cite{mcmahan2017communication}.

In an FL setup, user devices compute and transmit local model updates based on the local training data which are aggregated at the central server, facilitating users to learn collaboratively. With high-performance processors, modern mobile devices are equipped to handle such intensive computations, further aiding the implementation of FL in MEC. This has enabled its presence in a variety of delay-sensitive areas ranging from smart healthcare devices to predictive models from electronic health records. In particular, smart health applications have seen substantial success since they leverage the bulk data generated by tracking physical activities of its users from wearable devices such as smart watches, fitness trackers, and wristbands, to train quality learning models. Moreover, FL satisfies the privacy requirements of wearable computing by leaving personal data on the user devices~\cite{chen2020fedhealth}.

For these extensive time-critical applications, the feasible offloading time has to be in the order of milliseconds~\cite{al2018achieving}. In reality, without central availability of data, computing nodes need to communicate model updates often to attain convergence in FL. Communication of these updates may involve long round-trip times posing a limitation to this paradigm~\cite{trinh2018energy}. Moreover, unreliable and unpredictable network connections between the server and mobile devices could obstruct smooth transmission of updates. A large number of participants utilizing the constrained wireless bandwidth to upload model updates could add to uplink transmission delays. Therefore, given the nature of frequent exchange of updates in FL, over an expensive communication involving large number of mobile devices, reducing the overall time delay is crucial.

To address these challenges, many pioneering works analyze different aspects of the FL paradigm. Initial works such as~\cite{mcmahan2017communication} emphasizes on higher local computation to reduce the communication cost but lacks a theoretical model. Variants of FedAvg such as~\cite{konevcny2016federated} and works on distributed optimization such as~\cite{khaled2020tighter} aim to ease the communication burden. However, these works do not consider the limiting factors of wireless communication that can affect the performance of FL. Further, recent works including~\cite{tran2019federated} formulate to reduce the time or energy consumption but do not contemplate the learning hyperparameters which significantly affect the training time.

While majority of the works focus on communication overhead, the latest surge of research in networks have paved way for the rapidly expanding $5$G and the upcoming $6$G networks which alleviate communication burdens~\cite{shi2021towards}. To illustrate, a single-step of local computation on ResNet$50$ model over GPU consumes few hundreds of milliseconds~\cite{goyal2017accurate}, which is nearly comparable to the time taken to transmit over a wireless connection with transmission rate of $1$ Gbps. Therefore, it is worthwhile to investigate the impacts of communication, local computation in conjunction with convergence over FL. 

Intuitively, if a user performs more local computation to achieve a high local accuracy, frequent communication can be avoided due to decrease in the number of model updates. However, in case of data that is not representative of the overall distribution, this leads to local overfitting, adding to the convergence delay~\cite{rothchild2020fetchsgd}. On the contrary, to reduce computation, we can perform single-step updates which consumes lesser time to compute and communicate each update. However, it results in additional communications to update the current model, in order to attain a targeted global accuracy, increasing the overall time. As a result, the trade-off between wireless communication (to talk) and local computation (to work) of mobile devices needs to be balanced.

In this paper, we mainly aim to realize a balance between the two, by carefully studying the effect of various parameters, constraining overall time as the principal factor. We observe that for FL on each mobile device, the `talking' (i.e., global communication) time is determined by the local update size as well as wireless parameters such as transmission power, channel gain, bandwidth and background noise. Correspondingly, the `working' (i.e., local computation) time of each mobile device is influenced by the training data size and hyperparameters, together with the processor capabilities such as number of cycles and frequency scales. The overall time is further conditioned by the preset accuracies and the number of connected mobile devices. Capturing this motivation, our salient contributions can be summarized as follows,
\begin{itemize}
    \item
        We build a theoretical model for FL on edge GPUs over wireless networks that considers the impact of both computation and communication models on the overall time of training. To this end, we formulate an optimization problem to minimize the overall time consumed and reduce the number of communication rounds required to achieve FL convergence.
    \item
         Based on this model, we propose a delay-efficient FL solution mechanism by optimizing the influencing parameters to reduce the overall time. To realize this, we further consider the trade-off between local computation (to work) and wireless global communication (to talk). We demonstrate the theoretical convergence of the model and further define computational values based on leveraging the frequency of GPUs.
    \item
        We verify the effectiveness of our solution mechanism through extensive simulations over real-world datasets and illustrate the influence of each parameter on the overall time delay. We demonstrate that our solution significantly reduces the overall time in comparison with the baseline methods, while still achieving high accuracy.
\end{itemize}

\section{Delay-Efficient Federated Learning (DEFL) and Model Description}\label{Sys_Mod}
\subsection{Federated Learning over Mobile Edge Computing}
We consider an MEC-assisted FL system consisting of one edge (parameter) server and a set of $\mathcal{M}$ of $M$ mobile devices. Each mobile device $m$ has a local dataset $\mathcal{D}_m$ of size $D_m$, constituting a set of input samples and labels, $\{x_i^m, y_i^m\}_{i=1}^{D_m}$ with $d$ features. The loss function $F$ with respect to model parameters $\mathbf{w}$ on $m$'s dataset is given by, 
\begin{equation}
    \mathtt{F}_m(\mathbf{w})= \frac{1}{D_m}\sum_{i\in\mathcal{D}_m}\mathtt{f}_i(\mathbf{w}),
\end{equation}
where $\mathtt{f}_i(\mathbf{w}) = \mathtt{f}_i(\mathbf{w}; x_i^m, y_i^m)$ is the loss on data point $i$. The objective of minimizing the global loss is of the form,
\begin{equation} \label{eqn:loc_obj}
    \min_{\mathbf{w}\in\mathbb{R}^d} \mathtt{F}(\mathbf{w})= \sum_{m=1}^{M}\frac{D_m}{D}\mathtt{F}{_m}(\mathbf{w}),
\end{equation}
where $D = \sum_{m=1}^{M}D_m$ is the total data size. 

\subsection{Computation Model} \label{subsec:comp}
Typically, CPUs incur high computation costs~\cite{tran2019federated} and in contrast, with increased processing power and memory bandwidth, GPUs lower the computational costs. Furthermore, its massively parallel architecture can efficiently handle compute-intensive manipulations making it most suitable for high performance deep learning models. Hence in our work, we build a model for FL over edge GPUs whose frequency $f_m \in \mathbb{R}^m$, can be given as, 
\begin{equation} \label{eqn:fm}
    f_m = \frac{1}{a_s + \frac{a_c}{f_c} + \frac{a_M}{f_M}},
\end{equation}
where $a_s$, $a_c$ and $a_M$ are constants related to static, core frequency $f_c$ (including all of GPU's cores) and memory frequency $f_M$, respectively~\cite{abe2014power}. $G_m$ is the number of GPU cycles required for local computation by a mobile device and can be measured offline. We use mini-batch stochastic gradient descent (SGD) in which the computation is conditioned by the given batch size $b$. The local computation time taken to execute a single iteration of GPU-accelerated mini-batch SGD at the $m$-th mobile device can be given by,
\begin{equation}\label{eqn:Tcomp}
    T_{m}^{cp} = \frac{G{_m}b}{f_m}.
\end{equation}

The proposed model can also be used with CPUs or other processors where $f_m$ in \eqref{eqn:Tcomp} is replaced by the given processor's frequency value. Since GPUs are capable of parallel execution and process the whole-batch samples simultaneously~\cite{li2021talk}, in our work, we assume a synchronous model implying parallel local computation by mobile devices. Hence, the computation time during each communication round depends on the value of the slowest computation i.e., the highest time consumed by any mobile device given by, 
\begin{equation} \label{eqn:T_cp}
    T_{cp} = \underset{m}{\max}~ T_{m}^{cp}.
\end{equation}

\subsection{Communication Model}
The downlink bandwidth used by the server to broadcast the updated global model is much larger than the uplink bandwidth used by the mobile devices to transmit their local updates. Since this leads to a minimal downlink time versus uplink time~\cite{tran2019federated}, we consider only the uplink time as the communication time. Further, we assume that the local model update size $s$ to be fixed and the same for all mobile devices. Considering the transmission bandwidth $B$, transmission power of the mobile device $m$ as $p_m$, $h_m$ being the channel gain of the link between the mobile device and the server, $N_o$ the background noise, the communication time of one model update from each mobile device to the parameter server can be given by,
\begin{equation}
    T_{m}^{cm} = \frac{s}{B\log_{2}{(1 + \frac{p{_m}h{_m}}{N_o})}}.
\end{equation}

Assuming a synchronous model for communication, the communication time per communication round is given by,
\begin{equation} \label{eqn:T_cm}
    T_{cm} = \underset{m}{\max}~ T_{m}^{cm}.
\end{equation}

\subsection{Overall Time}
The total computation time per communication round depends on the number of local iterations $V$ and the overall time depends on communication together with computation time. Hence, the total time consumed by the system for one communication round can be defined as,
\begin{equation} \label{eqn:T}
    T = T_{cm} + V T_{cp}.
\end{equation}

\subsection{To Talk or To Work}
Both communication and computation-intensive networks can significantly benefit from reduced communication as communication is expensive. In addition, factors such as slow speed, poor communication channel, congestion in networks further challenge the efficient communication of model updates. Thus, reducing communication is a necessity in comparison with computation. In this aspect, when mobile devices perform more local computation to reach a high preset local accuracy, the number of local updates is reduced, indeed reducing the frequency of communicating with the server. This suggests fewer communication rounds implying savings in communication cost and time. Correspondingly, when functions across users share some similarity, taking local steps can lead to faster convergence~\cite{karimireddy2019scaffold}. Moreover, since the recent mobile edge devices are equipped with fast processors, increasing local computation does not burden or compromise the computation time. Adding parallel computing capabilities with the utilization of GPUs further aids in speeding up computation as described in Section \ref{subsec:comp}. Hence, we reduce the `talking' over `working' when balancing the trade-off.

\subsection{DEFL Algorithm}
Our methodology of FL named DEFL (\textbf{D}elay \textbf{E}fficient \textbf{F}ederated \textbf{L}earning), is described in Algorithm \ref{sec:algo}. The problem is formulated at the system-level and the computed values from the proceeding sections are utilized in our algorithm.
\addtolength{\topmargin}{0.01in}
\begin{algorithm}[ht]
\caption{DEFL}
\label{sec:algo}
\textbf{Inputs:} $\mathbf{w}_0$, preset global convergence error $\epsilon$, computed values of $b^*$ and $\theta^* \in [0, 1]$.
\begin{algorithmic}[1]
\STATE Initialize $\mathbf{w}_0$
\FOR{$1$ to $H$ communication rounds for achieving $\epsilon$,}
\STATE \textbf{Local Computation}: Each mobile device $m$ performs local training to compute stochastic gradient on mini-batch sized $b^*$, and solves \eqref{eqn:loc_obj} in $V$ local rounds to achieve $\theta ^*$-approximate solution.
\STATE \textbf{Wireless Communication}: Every participating mobile device $m$ transmits the local model update $w_v ^m$ to the edge server through the communication channel. 
\STATE \textbf{Aggregation and Broadcast}: The parameter server aggregates the received updates to obtain the global model, and broadcasts it to the mobile devices.
\ENDFOR
\end{algorithmic}
\end{algorithm}

\section{Theoretical and Convergence Analysis}
To present the theoretical analysis, we first state the following standard assumptions on the local loss function $\mathtt{F}_m$.
\begin{assumption}
The loss function $\mathtt{F}_m$ is $L$-smooth, that is for all $\mathbf{v}$ and $\mathbf{w}$, we have  $\mathtt{F}_m(\mathbf{v})\leq \mathtt{F}_m(\mathbf{w}) + (\mathbf{v}-\mathbf{w})^T\nabla \mathtt{F}_m +\frac{L}{2}\|\mathbf{v}-\mathbf{w}\|^2$.
\end{assumption}

\begin{assumption}
Let $\xi_k^m$ be sampled from the $m$-th device's local data uniformly at random. The variance of stochastic gradients in each device is bounded, i.e., $\mathbb{E}\|\nabla\mathtt{F}_m(\mathbf{w}_k^m,\xi_k^m)-\nabla\mathtt{F}_m(\mathbf{w}_k^m)\|\leq \sigma^2$.
\end{assumption}

The convergence bound of the model can be given by the following theorem using $\mathbf{w}_*$ as a fixed minimizer of $\mathtt{F}$.
\begin{theorem}[\cite{khaled2020tighter}]\label{ssgd}
Suppose Assumptions 1 and 2 hold, and a constant stepsize $\eta$ such that $\eta = \frac{\sqrt{M}}{4L\sqrt{K}}$ is chosen and the FL algorithm is run on identical data, then we have,
\begin{align*}\label{theo:1}
     \mathbb{E}\left[\mathtt{F}(\Bar{\mathbf{w}}_K) - \mathtt{F}(\mathbf{w}_*)\right] \leq& \frac{8\|\mathbf{w}_{0}-\mathbf{w}{_*}\|^2}{\sqrt{MK}} + \frac{\sigma^{2}}{2L\sqrt{MK}}  \\ &+\frac{\sigma^2{M}(V-1)}{LK}, \numberthis
\end{align*}
where $\Bar{\mathbf{w}}_K = \frac{1}{K}\sum_{k=1}^K\hat{\mathbf{w}}_{k}$ and $\hat{\mathbf{w}}_{k} = \frac{1}{M} \sum_{m=1}^{M}{\mathbf{w}}_{k}^{m}$. Additionally, the number of gradient steps is $K$, local rounds is $V$, and mobile devices is $M$.
\end{theorem}

\begin{remark}
The result of Theorem \ref{ssgd} is based on each user only computing a single stochastic gradient in each global iteration. However, in our FL setting, each mobile device computes a mini-batch of size $b$ in each communication round. Thus, we present the following corollary to show the convergence of DEFL.
\end{remark}

\begin{corollary}\label{cor:cor1}
Suppose Assumptions 1 and 2 hold, and a constant stepsize $\eta$ such that $\eta = \frac{\sqrt{M}}{4L\sqrt{K}}$ is chosen, with $K \geq M$ and the batch size equals $b$, then we have,
\begin{align*}\label{eqn:minibatch}
     \mathbb{E}\left[\mathtt{F}(\Bar{\mathbf{w}}_K) - \mathtt{F}(\mathbf{w}_*)\right] \leq& \frac{8\|\mathbf{w}_{0}-\mathbf{w}{_*}\|^2}{\sqrt{MK}} + \frac{\sigma^{2}}{2bL\sqrt{MK}}  \\ &+\frac{\sigma^2{M}(V-1)}{bLK}. \numberthis
\end{align*}
\begin{proof}
Mini-batch SGD is conditioned by the given batch size $b$. Using this in \eqref{theo:1}, we hence obtain this corollary.
\end{proof}
\end{corollary}

\begin{remark}\label{rem:b}
From Corollary \ref{cor:cor1}, we can observe that when each mobile device considers a mini-batch size $b$ in each iteration, it reduces the variance by a factor of $b$. 
\end{remark}

We now use the convergence properties of DEFL, to estimate the number of communication rounds required to complete training of the mobile devices in coordination with the edge server. We hence present the following corollary.
\begin{corollary}\label{cor:cor2}
The number of communication rounds for achieving an $\epsilon$-global model convergence, i.e, satisfying $\mathbb{E}\left[\mathtt{F}(\Bar{\mathbf{w}}_K) - \mathtt{F}(\mathbf{w}_*)\right] \leq \epsilon$ is given by,
\begin{align}\label{eqn:glob_it}
   H=\mathcal{O}\left(\frac{1}{b^{2}\epsilon^{2}MV} + \frac{M}{b\epsilon}\right),
\end{align}
where $\mathcal{O}$ is the big-$\mathcal{O}$ notation.
\begin{proof}
Since the system satisfies $\mathbb{E}\left[\mathtt{F}(\Bar{\mathbf{w}}_K) - \mathtt{F}(\mathbf{w}_*)\right] \leq \epsilon$ to achieve an $\epsilon$-accuracy, this is easily seen to be true by setting the right term in \eqref{eqn:minibatch} to $\epsilon$. Further, considering the relation of number of communication rounds, $H = K/V$ to solve for $H$ and using the big-$\mathcal{O}$ notation in \eqref{eqn:minibatch}, we thus obtain \eqref{eqn:glob_it}.
\end{proof}
\end{corollary}

\begin{remark}\label{remark:theta}
At the user level, for achieving a $\theta$-accuracy locally in SGD, i.e., $\mathbb{E}\|\mathbf{w}_{V}-\mathbf{w}_{*}\|_{2}^{2} \leq \theta$, the number of local rounds required for a mobile device's local model is $V = \nu \log \frac{1}{\theta}$~\cite{konevcny2017semi}, where $\nu$ is a constant related to step size and gradient noise. Then, substituting in \eqref{eqn:glob_it} and using the term $c$ to approximate the big-$\mathcal{O}$ notation we have,
\begin{align}\label{eqn_for_Q_with_theta}
    H = \frac{c}{b^{2}\epsilon^{2}M\nu\log\frac{1}{\theta}} + \frac{cM}{b\epsilon}.
\end{align}
\end{remark}
We can hence define the overall time for convergence as a product of the number of communication rounds required $H$, and the total time for one communication round $T$ as,
\begin{align}\label{eqn:T_o}
    \mathcal{T} = HT.
\end{align}

\section{Problem Formulation}
From our theoretical analysis, we can deduce the impact of batch size (shown in Remark \ref{rem:b}), number of communication rounds and time, preset accuracies and the number of participating mobile devices on the convergence rate. We hence achieve our objective of reducing the overall time by optimizing these variables. Accordingly, the optimization problem can be formulated using \eqref{eqn:T_o} with values from \eqref{eqn_for_Q_with_theta} and \eqref{eqn:T} as follows,
\begin{align}
    \underset{b, \theta, T_{cp}} {\operatorname{minimize}}&  \left(\frac{c}{b^{2}\epsilon^{2}M\nu\log\frac{1}{\theta}} + \frac{cM}{b\epsilon}\right) * \left(T_{cm} + \nu \log \frac{1}{\theta} T_{cp}\right) \label{PF: Expanded} \\
    {\mathrm{subject\ to}} &\quad
    b \in \{2^n  | n = {0, 1, ...}\} \label{PF:batch-size} \\
    &\quad0 \leq \theta \leq 1 \label{PF:theta} \\
   &\quad \underset{m}{\max}~ \frac{G{_m}b}{f_m} = T_{cp} \label{PF:tcp}
\end{align}

Constraint \eqref{PF:theta} defines the relative local accuracy that each mobile device attains on solving its local sub-problem. Here, $\theta=0$ corresponds to the exact solution and $\theta=1$ implies no improvement; hence we aim to achieve a lower value of $\theta$ for higher accuracy. This is also in accordance with \eqref{eqn_for_Q_with_theta}, which indicates that `working' more to achieve higher local accuracy results in smaller number of communication rounds. Although, this is in line with achieving our objective, \eqref{PF: Expanded} indicates that an inverse dependence on $\theta$ along with the relation with other parameters imply that we can only benefit a certain level by achieving a full relative accuracy of close to $0$. Hence, this control helps in avoiding local overfitting condition that otherwise delays convergence. Constraint \eqref{PF:batch-size} sets a range of the most commonly used effective batch size values starting from $1$, which is the case of SGD. For a given target global accuracy, a larger $b$ leads to smaller number of communication rounds as per \eqref{eqn_for_Q_with_theta}. Further, since we `work' more to achieve a preset local accuracy to balance the trade-off, computation time determined by the slowest computation is defined by constraint \eqref{PF:tcp}.

\section{Solution} \label{sec: solution}
The formulated problem to relieve the communication bottleneck by allowing more distributed computation is difficult to solve and involves a mix of integers and continuous variables. Hence, firstly, we introduce an auxiliary variable $\alpha = \log(1/\theta)$ to aid the optimization process, where $\alpha \in [0, +\infty)$ since $\theta \in [0, 1]$. Second, since constraint \eqref{PF:tcp} is non-convex, we can transform it to convex to alleviate solving. Third, we relax the constraint of $b$ in \eqref{PF:batch-size} from an integer to continuous; \eqref{PF: Expanded} can be reformulated as,
\begin{align}
    \underset{b, \alpha, T_{cp}} {\operatorname{minimize}}& ~~
    \left(\frac{c}{b^{2}\epsilon^{2}M\nu\alpha} + \frac{cM}{b\epsilon}\right) * \left(T_{cm} + \nu\alpha T_{cp}\right)\label{Sol} \\
    {\mathrm{subject\ to}} &\quad
    b \geq 1 \\
    &\quad\alpha \geq 0 \\
    &\quad T_{cp} \geq \frac{G{_m}b}{f_m},~~\forall{m \in \mathcal{M}}
\end{align}
\begin{proof}
We use Karush-Kuhn-Tucker (KKT) conditions to solve the delay minimization problem \eqref{Sol}. We first write the Lagrangian of \eqref{Sol} as follows,
\begin{multline} \label{sol2}
    \mathcal{L} (b, \alpha, T_{cp}, \lambda, \mu) = \left(\frac{cT_{cm}}{b^{2}\epsilon^{2}M\nu\alpha} + \frac{cMT_{cm}}{b\epsilon} + \frac{cT_{cp}}{b^{2}\epsilon^{2}M} \right.\\ + \left.\frac{cM\nu\alpha T_{cp}}{b\epsilon}\right) - \lambda{_1}(b - 1) - \lambda{_2}\alpha - \sum_{m=1}^M\mu{_m}\left(T_{cp} - \frac{G{_m}b}{f_m}\right), 
\end{multline}
where $\lambda_1$, $\lambda_2$, and $\{\mu{_m}\}_{m=1}^{M}$ are non-negative dual variables. 

We take the first order derivatives of \eqref{sol2} with respect to the dual and optimization variables giving the stationary conditions from \cref{eqn:dl/db,eqn:dl/da,eqn:dl/dTcp} and list the rest of the KKT conditions as in \cref{eqn:lam,eqn:mu,eqn:lam_mu} shown by,
\begin{align}\label{eqn:dl/db}
    \frac{\partial\mathcal{L}}{\partial b} &= \frac{-2cT_{cm}}{b^3 \epsilon^2 M \nu \alpha} - \frac{cT_{cm}M}{b^2 \epsilon} - \frac{2cT_{cp}}{b^3 \epsilon^2 M} \nonumber \\ 
    &- \frac{cMT_{cp}\nu\alpha}{b^2 \epsilon} - \lambda_1 + \frac{\mu_m G_m}{f_m} = 0,~~\forall{m \in \mathcal{M}},
\end{align}
\begin{align}\label{eqn:dl/da}
    \frac{\partial\mathcal{L}}{\partial \alpha} &= \frac{-cT_{cm}}{b^2 \epsilon^2 M \nu \alpha^2} + \frac{cT_{cm}M\nu}{b \epsilon} - \lambda_2 = 0,
\end{align}
\begin{align}\label{eqn:dl/dTcp}
    \frac{\partial\mathcal{L}}{\partial T_{cp}} &= \frac{c}{b^2 \epsilon^2 M} + \frac{cM\nu\alpha}{b \epsilon} - \mu_m = 0,~~\forall{m \in \mathcal{M}},
\end{align}
\begin{align}\label{eqn:lam}
    \lambda_1 (b-1)=0,~~\lambda_2 (\alpha)=0,
\end{align}
\begin{align}\label{eqn:mu}
    \mu_m \left(T_{cp}-\frac{G_m b}{f_m}\right)=0,~~\forall{m \in \mathcal{M}},
\end{align}
\begin{align}\label{eqn:lam_mu}
    \lambda_1 \geq0,~~\lambda_2 \geq0,~~\mu_m \geq0,~~\forall{m \in \mathcal{M}}.
\end{align}

Since the inequality constraints are nonlinear yet differentiable and lower-bounded with a non-negative duality gap, the KKT necessary conditions serve as the optimality conditions. Hence, considering the above dual feasibility and complementary slackness conditions to solve the derivatives, KKT points are obtained. We check all of the obtained points for feasibility of the problem to finally deduce the optimal values as,
\begin{align}
\begin{cases}  \label{E:Lemm1_a_2}
    &\alpha^* = \sqrt{\frac{T_{cm}f_m}{ M^2 \epsilon \nu^2 G_m}},~~\forall{m \in \mathcal{M}};\\
	&b^* = 2cM\sqrt{\frac{T_{cm}f_m \epsilon}{G_{m}}},~~\forall{m \in \mathcal{M}};\\
    &T_{cp}^*  = \max_{m}\frac{G{_m}b^*}{f_m},~~\forall{m \in \mathcal{M}}.
\end{cases}
\end{align}
\end{proof}

From these relations, theoretically, the computation time is vastly affected by loads from all the mobile devices and the processors' computational capabilities and speed. Further, the batch size has a direct impact on $T_{cp}$ with larger $b$ leading to higher computation and faster convergence. Both $b$ and the relative local error $\theta$ are impacted by the set global convergence error $\epsilon$, $M$, along with other parameters. A lower value of $\theta^*$ (which can be computed from $\alpha^*$) implying higher local accuracy, results in more `working' and less `talking'.
\begin{figure*}
    \centering
    \subfigure[Impact of $\epsilon$ on $\mathcal{T}$ and convergence.\label{fig:eps}]
    {\includegraphics[width=1.75in]{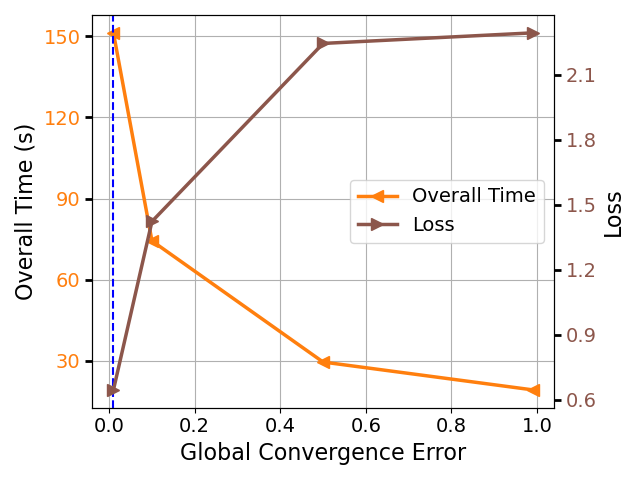}}
    \subfigure[Impact of $b$ with the chosen $\epsilon$.\label{fig:batch}]
    {\includegraphics[width=1.75in]{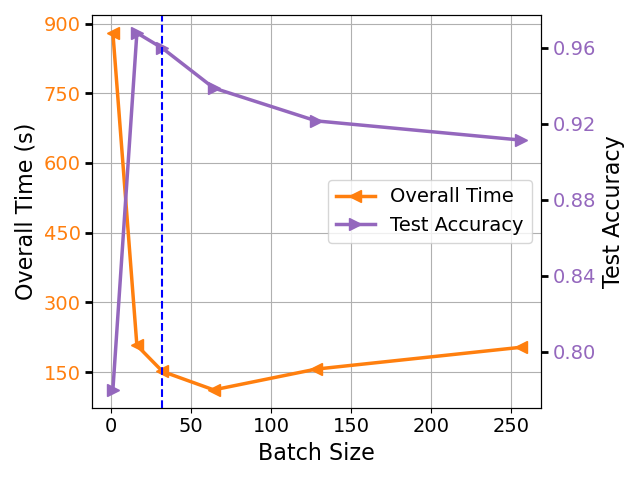}} 
    \subfigure[Impact of $\theta$ with $b^*$ and chosen $\epsilon$.\label{fig:theta2}]
    {\includegraphics[width=1.75in]{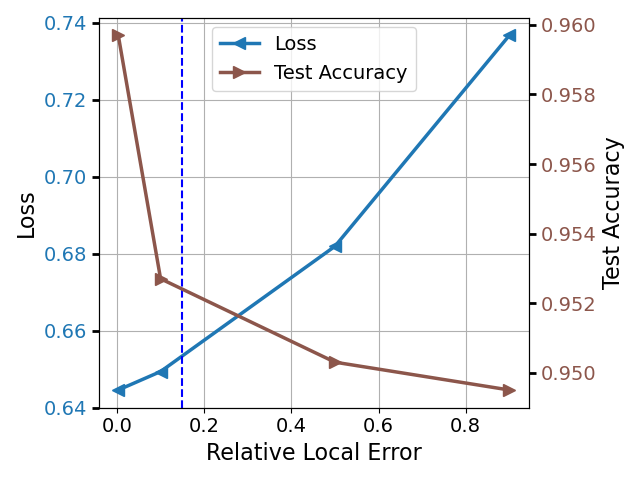}} 
    \subfigure[Impact of $\theta$, $T_{cp}$ on $H$.\label{fig:theta}]
    {\includegraphics[width=1.75in]{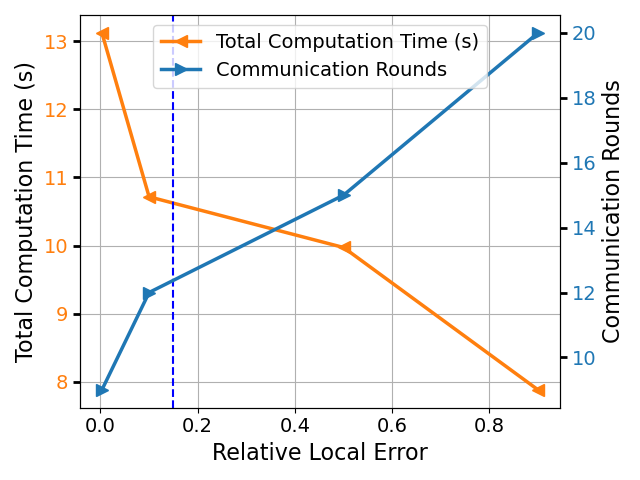}} 
    \caption{Studying the impact of different parameters on the overall time and performance.}
    \label{fig:1}
\end{figure*}

\section{Performance Evaluation}\label{sec:expt}
\subsection{Settings}
To evaluate the proposed delay efficient FL, we perform simulations using image classification tasks on the widely used MNIST\footnote{Downloaded from: \url{http://yann.lecun.com/exdb/mnist}} 
and CIFAR-$10$\footnote{Downloaded from: \url{http://www.cs.toronto.edu/~kriz/cifar.html}} datasets using CNN. For the FL tasks, we consider $1$ parameter server and $10$ mobile devices with distributed data and a learning rate of $0.01$. In accordance with our computational model in \eqref{eqn:fm}, we use Nvidia RTX$8000$ with the number of GPU cycles of $30$ cycles/bit and following constraint \eqref{PF:tcp}, we consider an equal maximum computation capacity of $f_m = 2$ GHz for all the mobile devices. For communication model, we assume the bandwidth $B=20$ MHz and noise $N_o = -174$ dBm/Hz.
\begin{figure*} 
    \centering
    \subfigure[Overall time vs. comm. rounds.\label{fig:mnist_time}]
    {\includegraphics[width=1.75in]{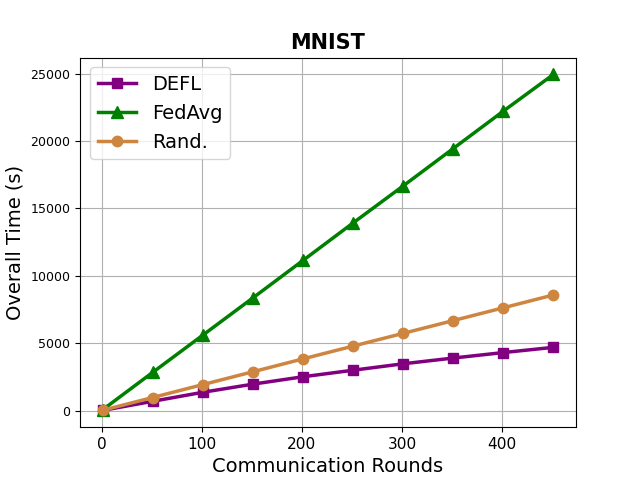}} \hfill
    \subfigure[Test accuracy vs. comm. rounds.\label{fig:mnist_acc}]
    {\includegraphics[width=1.75in]{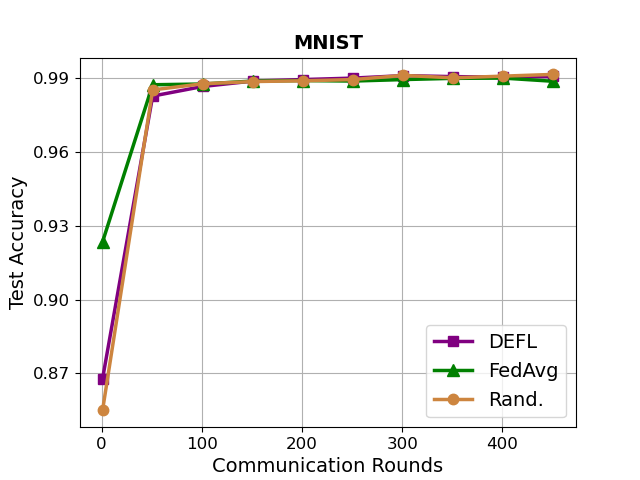}} \hfill
    \subfigure[Overall time vs. comm. rounds.\label{fig:cifar_time}]
    {\includegraphics[width=1.75in]{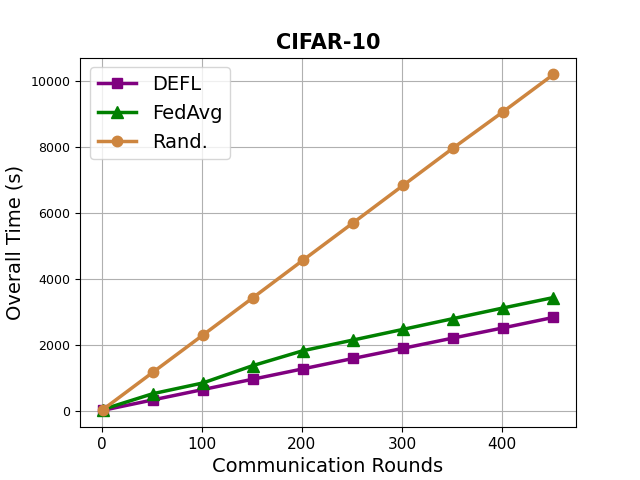}} \hfill
    \subfigure[Test accuracy vs. comm. rounds.\label{fig:cifar_acc}]
    {\includegraphics[width=1.75in]{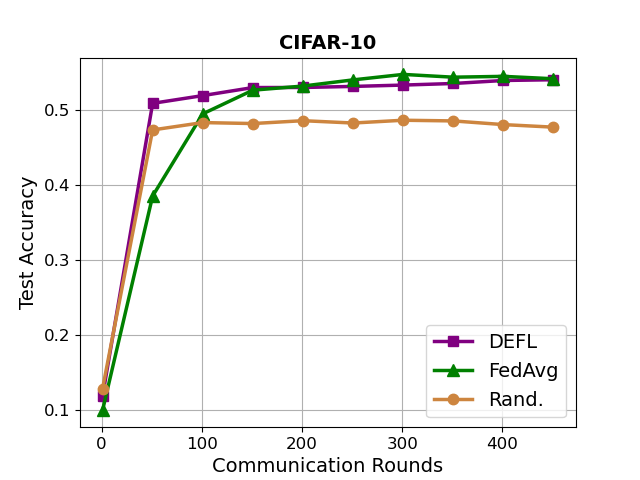}} 
    \caption{Performance evaluation of DEFL over MNIST and CIFAR-10 datasets.}\label{fig:2}
\end{figure*}

\subsection{Impact of optimization parameters over convergence}
According to (\ref{E:Lemm1_a_2}), the computed values of $b^*$, $\theta^*$ and in turn $T_{cp}^*$ are conditioned by the relative global convergence error $\epsilon$. Hence, we empirically choose a value which leads to both increased performance yet takes less overall time. From the values in Fig. \ref{fig:eps}, we thus set $\epsilon=0.01$. The optimized variables computed from our solution are used in \eqref{eqn_for_Q_with_theta} to determine the number of communication rounds $H$, which can be empirically shown as in Fig. \ref{fig:theta}. We now study the impact of the parameters on the overall time as follows.

\textbf{Batch size}.
Generally, larger batch size to train the model allows computational speedups from the parallelism of GPUs. However, too large a batch size may lead to lower generalization, resulting in more overall time. Whereas, smaller batch sizes are shown to have less computation but are not guaranteed to converge to the global optima. Theoretically, the value of $b$ computed from \eqref{E:Lemm1_a_2} has a lower limit of $1$ and can be rounded off to $32$ (for MNIST data size) which also corresponds to a value from the initial constraint (\ref{PF:batch-size}). Empirically, as shown in Fig. \ref{fig:batch}, to achieve the same target $\epsilon$, while $b = 64$ has the shortest overall time, it has a lower test accuracy. On the other hand, $b=16$ achieves the highest test accuracy but takes more time of about $200$ seconds. Consequently, the computed value of $b=32$ achieves a good trade-off between prediction performance and overall time. 

\textbf{Relative Local Error}.
A lower value of relative local error $\theta$ (i.e., higher local accuracy), induces the model to `work' more to achieve $\theta$-accurate solution locally. This implies that fewer communication rounds is necessary according to \eqref{eqn_for_Q_with_theta} and consequently, lesser communication time than the original FedAvg algorithm. This behavior is captured in Fig. \ref{fig:theta}, where the theoretically calculated $\theta \approx 0.15$ from \eqref{E:Lemm1_a_2} has a higher computation time (due to `working' more), but smaller $H$ due to reduced number of model updates. Conversely, higher $\theta$ is undesired since lower computation results in `talking' more with larger number of $H$ and higher overall time. Further, as shown in Fig. \ref{fig:theta2}, $\theta$ is just as low as to achieve a better performance in terms of reduced training loss at the same overall time while avoiding local overfitting.

\textbf{Computation Time}.
The computed batch size influences the computation time since the training dataset is processed batch-wise, subject to device capabilities. Accordingly, increasing $b$ implies taking advantage of the available computational resources of the mobile devices. As seen in Fig. \ref{fig:theta}, higher computation leads to reduced number of communication rounds which in turn leads to reduced overall time.

\textbf{Comparison with Baseline}.
For evaluation, we use Federated Averaging (FedAvg) from~\cite{mcmahan2017communication} as a baseline to compare the performance of our proposed solution. For FedAvg on MNIST IID data using CNN, we set the parameter values as recommended by the authors through their experiments as $b=10$ and $V=20$. We then choose random values of $b=16$ and $V=15$ for MNIST and $b=64$ and $V=30$ for CIFAR-10 to test the effect of parameters as a whole, marked by `Rand.'. For our work marked as `DEFL', we choose values as per our delay-efficient optimized solution from Section \ref{sec: solution} and as verified in Section \ref{sec:expt}. With a preset $\theta$ ensuring more computation, along with the optimized $b$ and fixed $\epsilon$, we observe from Fig. \ref{fig:2} that, although we achieve nearly the same test accuracy, DEFL significantly outperforms the baseline in terms of the overall time. Comparatively, we reduce the overall time by nearly $70\%$ compared with FedAvg for MNIST and by $18\%$ for CIFAR. Similarly, there is a reduction of around $38\%$ comparing with `Rand.' for MNIST and $75\%$ for CIFAR. Hence, DEFL can be useful in accelerating the FL process on mobile edge devices such as wearable devices.

\section{Conclusion}
In this paper, we introduced a delay efficient FL mechanism suitable for mobile edge devices such as wearable devices, by studying the trade-off between wireless communication (to talk) and local computation (to work) with respect to the overall time. With careful consideration of this prevailing balance, we interpreted the effects of the learning model, wireless communication and hyper parameters in conjunction over the total time consumed. Guided by this theoretical model, we demonstrated the impact of these parameters through extensive simulations. Empirical evaluations have shown that DEFL can reduce the overall time delay while achieving high performance accuracy, implying that FL can be accommodated in delay-sensitive applications suitable for mobile devices.
\bibliography{./pav}
\bibliographystyle{IEEETran}
\end{document}